\documentclass[10pt,twocolumn,letterpaper]{article}

 \usepackage[pagenumbers]{cvpr} 

\definecolor{cvprblue}{rgb}{0.21,0.49,0.74}
\usepackage[pagebackref,breaklinks,colorlinks,allcolors=cvprblue]{hyperref}

\usepackage{xcolor}

\def\paperID{17942} 
\def\confName{CVPR}
\def\confYear{2026}

\title{Selecting Fine-Tuning Examples by Quizzing VLMs}

\author{Tenghao Ji\\
University of Michigan\\
{\tt\small jimji@umich.edu}
\and
Eytan Adar\\
University of Michigan\\
{\tt\small eadar@umich.edu}
}

\usepackage{listings}
\begin{document}



\maketitle
\begin{abstract}
A challenge in fine-tuning text-to-image diffusion models for specific topics is to select good examples. Fine-tuning from image sets of varying quality, such as Wikipedia Commons, will often produce poor output. However, training images that \textit{do} exemplify the target concept (e.g., a \textit{female Mountain Bluebird}) help ensure that the generated images are similarly representative (e.g., have the prototypical blue-wings and gray chest). In this work, we propose QZLoRA, a framework to select images for low-rank adaptation (LoRA). The approach leverages QuizRank, a method to automatically rank images by treating them as an `educational intervention' and `quizzing' a VLM. We demonstrate that QZLoRA can produce better aligned, photorealistic images with fewer samples. We also show that these fine-tuned models can produce stylized that are similarly representative (i.e., illustrations). Our results highlight the promise of combining automated visual reasoning with parameter-efficient fine-tuning for topic-adaptive generative modeling.
\end{abstract}


\section{Introduction}

Text-to-image diffusion models such as Stable Diffusion~\cite{rombach2022high}, 
DALL-E~\cite{ramesh2022hierarchical}, and Imagen~\cite{saharia2022photorealistic} can produce incredibly high-quality images that align with natural language prompts. However, despite their versatility, diffusion models still suffer from uneven generation quality across topics. In particular, some subjects may not display characteristic visual properties that are distinguishing features. For example, there are subtle differences between a Mountain, Western, and Eastern Bluebird (varying chest colors) that many models simply blend together. his performance gap stems from both the distributional bias of the pretraining data and the lack of fine-grained topic adaptation.

\begin{figure}[t]
  \centering
  \includegraphics[width=\linewidth]{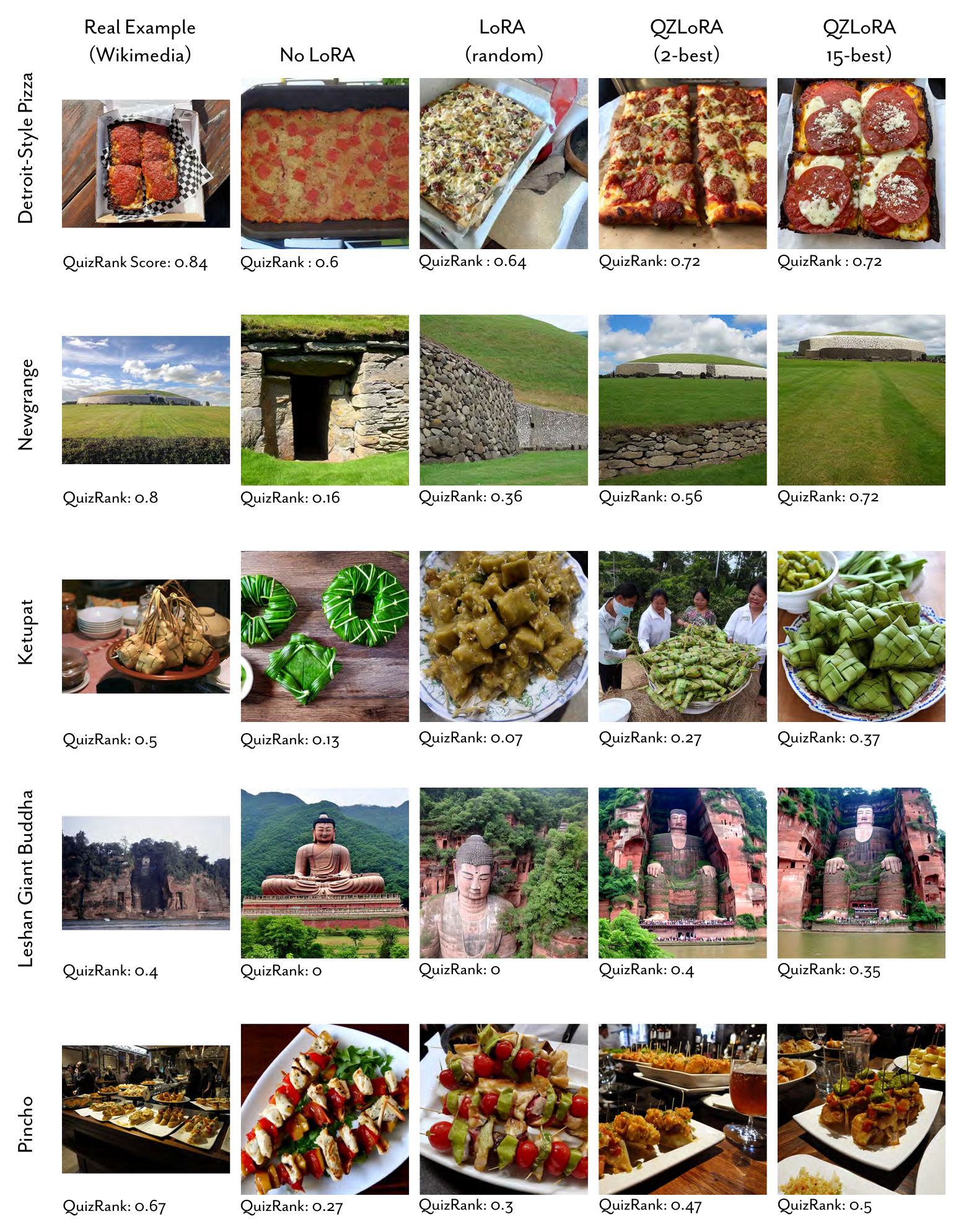}
  \caption{Example outputs for various topics (in rows). From left to right: a real example from the Wikimedia Commons, a generated photorealistic image with no fine-tuning, an image generated with a LoRA using a random sample of Commons images, and a LoRA model tuned with the 2 best QuizRank scored images, and an image from the LoRA tuned on the 15 best images. The QuizRank score is displayed below each image.}
  \label{fig:teaser_realistic}
\end{figure}

To mitigate such biases, parameter-efficient fine-tuning methods such as Low-Rank Adaptation (LoRA)~\cite{hu2022lora} have been widely adopted. LoRA enables lightweight adaptation of large diffusion models to specific domains or styles using relatively small training sets, while preserving the general visual priors of the base model. However, the effectiveness of LoRA fine-tuning still depends on the quality of the selected training data: random or noisy samples can lead to overfitting or degraded visual coherence (as can be seen in Figures~\ref{fig:teaser_realistic} and~\ref{fig:teaser_illustration}). Popular images (e.g., those most commonly used in Wikipedia) may not be popular because they are good~\cite{ji2025quizrankpickingimagesquizzing} but due to diffusion pressures~\cite{he2018the_tower_of_babel}. This raises a central question—\emph{how can we automatically identify the most representative and reliable images for LoRA training?} 

To address this challenge, we build on the \emph{QuizRank} algorithm~\cite{ji2025quizrankpickingimagesquizzing}. QuizRank works by treating images an an `educational intervention' for Vision-Language Models (VLMs). It first generates a `test' using distinguishing visual characteristics of the target class. This is done by analyzing the textual description of the object in a source such as Wikipedia to generate multiple choice questions. If the VLM can answer more questions correctly when `seeing' image A than image B, image A will gain a higher QuizRank score. This approach has two critical benefits for LoRA. First, it allows us to rank images based on alignment to visually important characteristics of target objects (i.e., images that \textit{exemplify} the target class). The ranking of images in this way allows for the selection of a small but effective training set for fine-tuning. Second, the algorithm allows us to evaluate the outputs of diffusion algorithms to evaluate their quality objectively. Models that produce images that are better visually-aligned with the target class will have a higher QuizRank score.

In this paper, we integrate QuizRank with LoRA fine-tuning to automatically select the best-performing images. We demonstrate how this QZLoRA pipeline can lead to improved outputs of Stable Diffusion across a wide array of topics. Through systematic experiments, we demonstrate that QuizRank-guided LoRA fine-tuning produces significantly better-aligned images than various baselines (i.e., untuned or tuned with random in-class images). We further employ QuizRank itself as an objective evaluation metric, measuring the question-answering accuracy of generated images to assess both their visual quality and semantic correctness. In addition to photorealistic generated images, we show how our fine tuning can also produce better representative images in other styles (e.g., illustrations). These results confirm the robustness of our approach across many visual domains.

\section{Related Work}

\subsection{Challenges in Text-to-Image Diffusion}

Although diffusion models have achieved impressive progress, their performance remains uneven across topics. Generic subjects (e.g., `a bird') are often rendered with high fidelity, whereas highly specific (e.g., `a Mountain Bluebird'), rare, or abstract topics tend to produce blurred or visually inaccurate results. This discrepancy is closely tied to the long-tailed distributions of large-scale pretraining datasets, where frequent visual categories dominate representation learning. Prior studies have shown that such imbalance can lead to systematic topic and cultural bias~\cite{birhane2021multimodal, luccioni2023stable}, resulting in uneven coverage of visual concepts and diminished diversity in generated outputs.

Recent work has also revealed that these biases can be amplified during text-to-image generation, influencing how concepts are visualized in social and cultural contexts~\cite{schneider2025investigating, seshadri2023bias}.
In particular, long-tailed data distributions can degrade diffusion models' performance on rare or fine-grained concepts~\cite{zhang2024long}, while cross-cultural evaluations reveal substantial performance disparities across visual domains from different regions~\cite{rege2025cure}.

\subsection{Parameter-Efficient Fine-Tuning}
Parameter‑efficient fine‑tuning (PEFT) enables large generative models to adapt to new domains without retraining all parameters. Methods like prompt tuning~\cite{lester2021power} and low‑rank adaptation (LoRA)~\cite{hu2022lora} add lightweight, trainable components to the model, preserving its general knowledge while focusing on the new task. LoRA specifically introduces low‑rank matrices to existing layers, enabling efficient fine‑tuning. Marjit et al. (2024) propose a Kronecker‑product based adapter for diffusion models, reducing the number of trainable parameters while improving fidelity~\cite{marjit2025diffusekrona}. Tao et al. (2025) extend LoRA with tensor decomposition for improved adaptation~\cite{tao2025transformed}.

PEFT has been successfully applied to tasks such as style transfer and subject‑driven generation. These approaches have been popularized by methods such as DreamBooth~\cite{ruiz2023dreambooth} and Textual Inversion~\cite{gal2022image}. However, their performance heavily depends on the quality of the fine‑tuning data. Poorly selected or noisy images can degrade the output. This highlights the need for automated methods for data selection~\cite{moon2022fine}. Our proposed method presents an approach to optimize data selection by identifying representative images of a topic/concept.

\subsection{Automated Image Evaluation}
Recent literature has increasingly focused on dedicated evaluation frameworks for text-to-image generative models. For example, Hartwig et al. provide a comprehensive survey of quality metrics for text-to-image generation~\cite{hartwig2025survey}, while Chen et al. introduce an empirical study assessing both aesthetic qualities and semantic-concept coverage~\cite{chen2024evaluating}. These approaches underscore the need for robust evaluation methods that capture both the perceptual and conceptual fidelity of generated images.

Automatic assessment of generated images is often based on similarity or preference metrics such as CLIPScore~\cite{hessel2021clipscore} and Human Preference Score~\cite{wu2023human}. Although these measures reflect global alignment or perceptual appeal, they are limited in their ability to assess whether an image truly conveys the intended concept. To address these limitations, recent studies have explored vision-language reasoning and visual question answering (VQA) for a more semantic evaluation. For example, TIFA generates question-answer pairs to assess how well a generated image matches its textual prompt~\cite{hu2023tifa}. However, many of these approaches rely on handcrafted questions or small-scale benchmarks that restrict their scalability and applicability across diverse domains. Our approach uses QuizRank~\cite{ji2025quizrankpickingimagesquizzing}, an alternative approach to generate questions and evaluate images through the use of VLMs.

\section{Design}

\begin{figure*}[t]
    \centering
    \includegraphics[width=\textwidth]{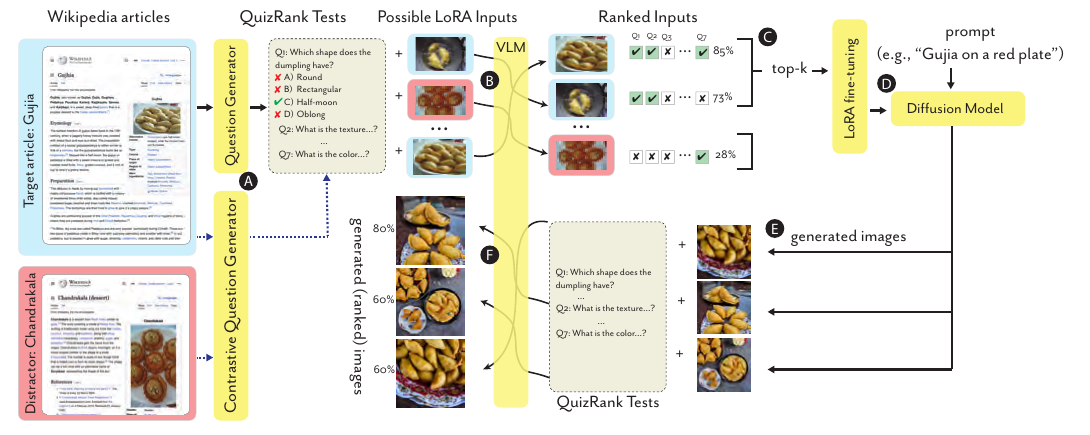}
    \caption{The QZLoRA approach: A test is generated based on visual properties of the target object by using textual descriptions of the target and distractors (step A). Each possible image and test are fed into a VLM (step B). Images are ranked based on how many questions the VLM answers correctly (step C) and the top-k are fed to LoRA for fine-tuning (step D). A Stable Diffusion model is used to generate new images (step E). Optionally, these images can be ranked (step F) using the original QuizRank test (elements recreated from~\cite{ji2025quizrankpickingimagesquizzing})}
    \label{fig:flow}
\end{figure*}

The QZLoRA procedure proceeds in a few steps (depicted graphically in Figure~\ref{fig:flow}). We utilize the QuizRank algorithm~\cite{ji2025quizrankpickingimagesquizzing} (step A) to produce a `test' for the images. The approach uses text descriptions of the target class (e.g., from a Wikipedia page) to produce a set of multiple-choice questions using an LLM. A variant of this approach (depicted as the `contrastive question generator' in the diagram) will also contrast the target class to similar distractor classes. For example, the Gujia pastry is very similar to the Chandrakala. This allows QuizRank to produce questions that specifically target the unique visual characteristics of the target class (e.g., Gujia). For example, questions may focus on the texture, material, size, shape, or other distinguishing characteristics of the target class. The intuition is that if the image works well as a `graphical aid' to the VLM, the VLM will do better with the test.

Once we have constructed this test, it is possible to rank a set of possible inputs to the LoRA training. Images can include both real instances of the target object (e.g., as found in Wikimedia Commons) and distractor images for validation. Each image and test are passed to the VLM for evaluation (step B). From here, images can be ranked on the basis of how well they perform. The complete prompts used for this procedure are provided in~\cite{ji2025quizrankpickingimagesquizzing}. The approach has been shown to be highly correlated with human performance. That is, the VLM can approximate how a human looking at the image and answering the questions would perform. As results have been shown to be roughly stable on closed and open models, we used OpenAI's GPT-4o model\footnote{Specifically, \texttt{gpt-4o-2024-11-20}} as both LLM and VLM.

For training the LoRA model (step C), we select the top-$k$ ranked images. For evaluation purposes, $k$ ranged from 2 to 15 images. To train these models, we used the open-source \texttt{kohya\_ss} framework, which provides a convenient interface for fine-tuning diffusion models. All LoRA models were trained in identical hyperparameter settings: \texttt{epochs=20}, \texttt{num\_repeats=5}, \texttt{batch\_size=1}, \texttt{learning\_rate=1e-4}, \texttt{AdamW8bit} optimizer, and a maximum resolution of \texttt{512×512}. The computational cost for training was significant, and all fine-tuning tasks were performed on an NVIDIA GeForce RTX 4060 Laptop GPU with 8 GB of memory. Generating a fine-tuned model with 15 images took approximately 13 minutes. The training time varied depending on the number of images used, with fewer images leading to slightly faster training times.

The fine-tuned LoRA can then be used in conjunction with other text-to-image models (step D). In our implementation, we utilize the Stable Diffusion 1.5 model\footnote{The \texttt{v1-5-pruned-emaonly} checkpoint (SHA: 6ce0161689)}. We are then able to generate new images (photorealistic or otherwise) using standard prompting (example prompts are provided in the appendix). The generated images (step E) can be optionally evaluated using the previously generated QuizRank test. This allows for additional tuning of the LoRA model or selecting higher quality generated outputs among a set.

\section{Evaluation}
To evaluate the QZLoRA approach, we contrast the performance of different generative procedures (both fine-tuned and not). The advantage of utilizing QuizRank--in addition to determining which images to use in fine-tuning--is that it will allow us to objectively compare the output images.

We utilize the following generative procedures for comparison. As we describe in detail below, each target topic has a large set, $S_{topic}$, of possible images that can be used for LoRA fine tuning.
\begin{itemize}
    \item \textbf{No-LoRA baseline}, images are generated with no LoRA fine-tuning using the base Stable Diffusion~1.5 model
    \item \textbf{LoRA Random-15}, Fifteen random items are selected from $S_{topic}$ and used to train the LoRA. This model is then used in conjunction with the diffusion model to generate images.
    \item \textbf{QZLoRA top-2}, The top 2 images from $S_{topic}$ based on QuizRank scores are used to train the LoRA.
    \item \textbf{QZLoRA top-15}, The top 15 images (again, based on QuizRank) are used for fine-tuning.
\end{itemize}

In our experiment, we selected 60 topics for analysis. These topics were selected according to the procedure described by Silva et al.~\cite{silva2024imagine}. 
To ensure generalizability, topics were selected from the Biology, Architecture, Food \& Drink, and Art categories. To keep a somewhat uniform level of `difficulty', the chosen topics were active (i.e., were edited and visited) but were not among the most popular (under 6000 views per month). To mitigate data scarcity bias, topics were also selected to ensure that there were at least 30 publicly available images on Wikimedia Commons.

The entire pipeline—from image collection to LoRA training and image generation—was fully automated. Given a \texttt{topic url} and a short textual label (typically the first sentence from the corresponding Wikipedia article), the system downloads up to 55 images from the Wikipedia Commons (each paired with a caption file of matching index). QuizRank was applied to the topic's Wikipedia page to generate questions, and then each of the images was evaluated to produce a ranked output. 

\subsection{Generation Procedure}
To evaluate the generative process, we used a summary sentence of the topic (derived from Wikipedia) as the positive prompt for that topic. To reduce the effect of stochasticity in image generation, each condition produced five images per topic, and the average performance of these samples was used for comparison. To control stylistic bias, we applied targeted negative prompts. For photorealistic images, we also provided a negative prompt (as is standard in many generative applications) to exclude artistic and low-quality terms (e.g., illustration, cartoon, painting).

\begin{figure}[t]
  \centering
  \includegraphics[width=\linewidth]{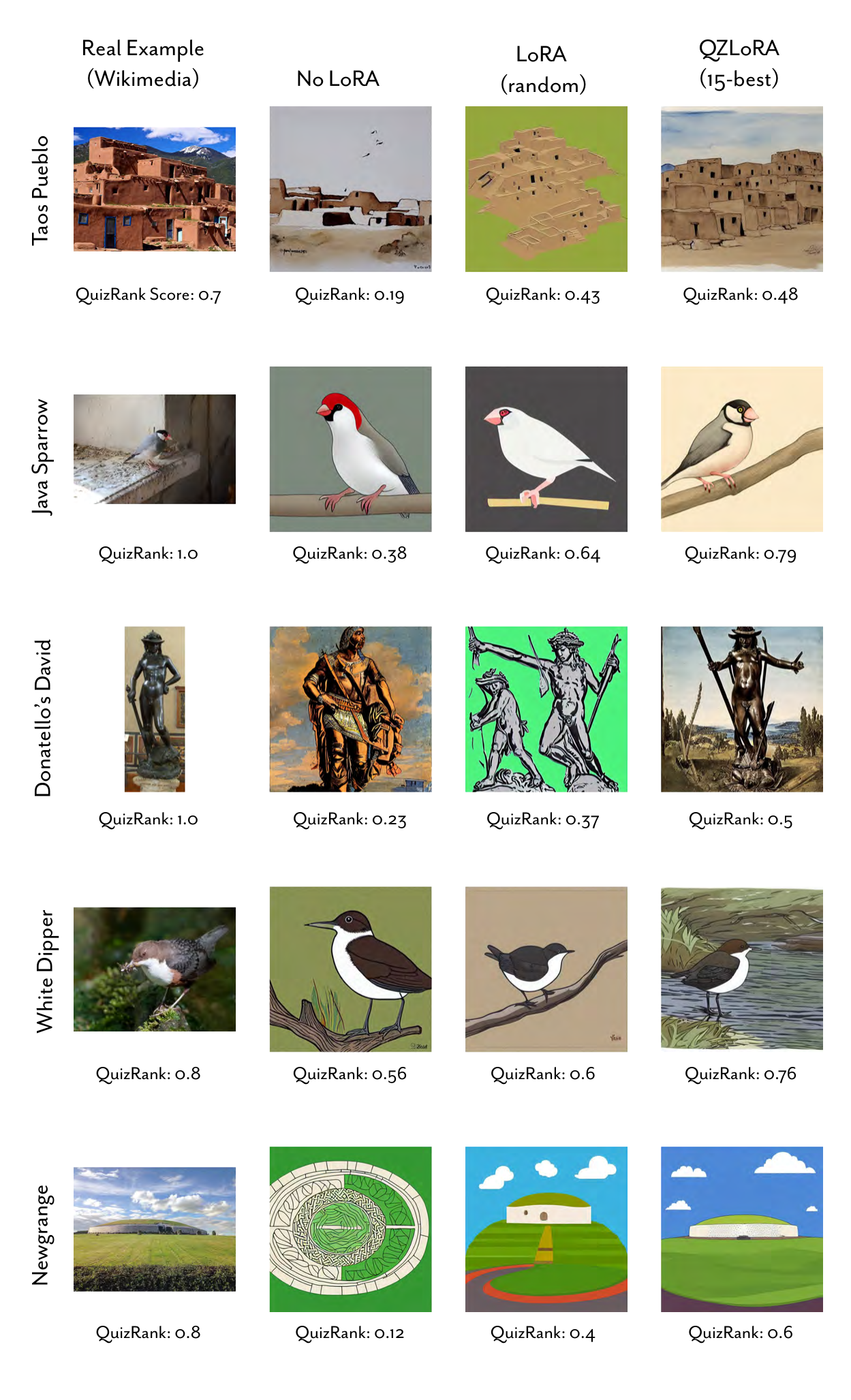}
  \caption{Example outputs for various topics (in rows). From left to right: a real example from the Wikimedia Commons, a generated illustrated image with no fine-tuning, an image generated with a LoRA using a random sample of Commons images, and a LoRA model tuned with the 15 best QuizRank scored images. The QuizRank score for each image is displayed below each image.}
  \label{fig:teaser_illustration}
\end{figure}

To examine whether QZLoRA benefits beyond photorealistic generation to more stylized visual domains, we conducted a supplementary \textbf{illustration-style} evaluation (see Figure~\ref{fig:teaser_illustration} for examples). This process was evaluated using the `No LoRa' baseline and `QZLoRA top-15.' For illustration-style images, photorealistic terms were used as part of the negative prompt (e.g., realistic, photo, CGI).

For each topic and condition, the five generated images were evaluated using QuizRank (with the original test produced for that topic). The final accuracy per topic was calculated as the average QuizRank score on the five generated images. See the Appendix for example generative prompts.

\subsection{Overall Performance Comparison}

\begin{figure}[t]
\centering
\includegraphics[width=\linewidth]{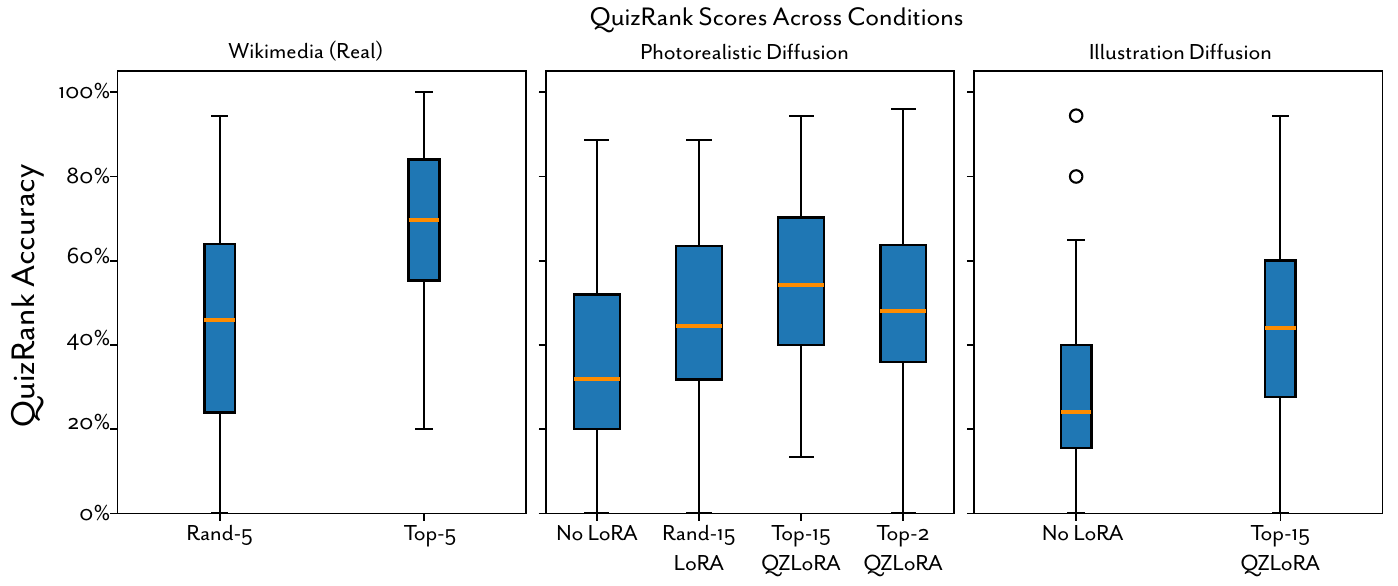}
\caption{Boxplot of QuizRank accuracy under different conditions. LoRA fine-tuning guided by QuizRank-selected images shows higher median accuracy and reduced variance compared to random or baseline settings.}
\label{fig:boxplot}
\end{figure}

Figure~\ref{fig:boxplot} illustrates the distribution of QuizRank accuracy for the six experimental conditions (four photorealistic and two illustration). In addition, the first two columns report on the QuizRank scores for the topic for \textit{real} images of that object. This is done for five random (\textbf{rand-5}) images from $S_{topic}$ and \textbf{top-5} (based on QuizRank score). This provides a sense of a ceiling score for images that already exist in Wikipedia (i.e., are not generated). We note that many topics do not have images that achieve a perfect QuizRank score and we return to this in the discussion.  Our results show a consistent pattern: LoRA fine-tuning based on QuizRank-selected images tend to produce higher and more stable generation quality than the random or baseline settings. More specifically:

\textbf{(1) Central trend.}  
Among the realistic-style conditions, the model fine-tuned on the top-15 QuizRank images reaches the highest average accuracy (\textbf{54.99\%}) and median (\textbf{54.34\%}).  
Both values are notably higher than those of random selection (\textbf{46.74\%}) and the no-LoRA baseline (\textbf{35.37\%}).  Even using only the two highest ranked images (\textbf{50.03\%}), performance remains above the random case, suggesting that the quality of training examples has a stronger effect than quantity.

\textbf{(2) Stability across topics.}  
The dispersion of the results is smaller for QuizRank-guided fine-tuning (\textbf{std = 19.04\%}) compared to random sampling (\textbf{21.37\%}). Although the difference is small, it suggests that image selection based on visual–semantic accuracy helps the model generalize more evenly across different topics.

\textbf{(3) Outliers and difficult topics.}  
Some topics (which we discuss more below) still show low scores even after adaptation, which likely corresponds to abstract or visually ambiguous subjects. In contrast, several topics approach near-perfect accuracy (\textbf{$>$90\%}) after fine-tuning, showing that QuizRank can identify highly representative samples when sufficient visual cues are available.

\textbf{(4) Real and illustration domains.}  
The real images from Wikipedia serve as a reference point (\textbf{69.18\%} on average). After fine-tuning, realistic generations approach this range, while illustration-style results remain lower in absolute value (\textbf{43.62\%} vs.\ \textbf{28.49\%} for the baseline) but follow the same trend of improvement. In general, QuizRank-based sample selection improves both realism and topic relevance, with clearer benefits in realistic image synthesis.

\subsection{Topic-Level Comparison} 

\begin{figure}[t]
  \centering
  \includegraphics[width=\linewidth]{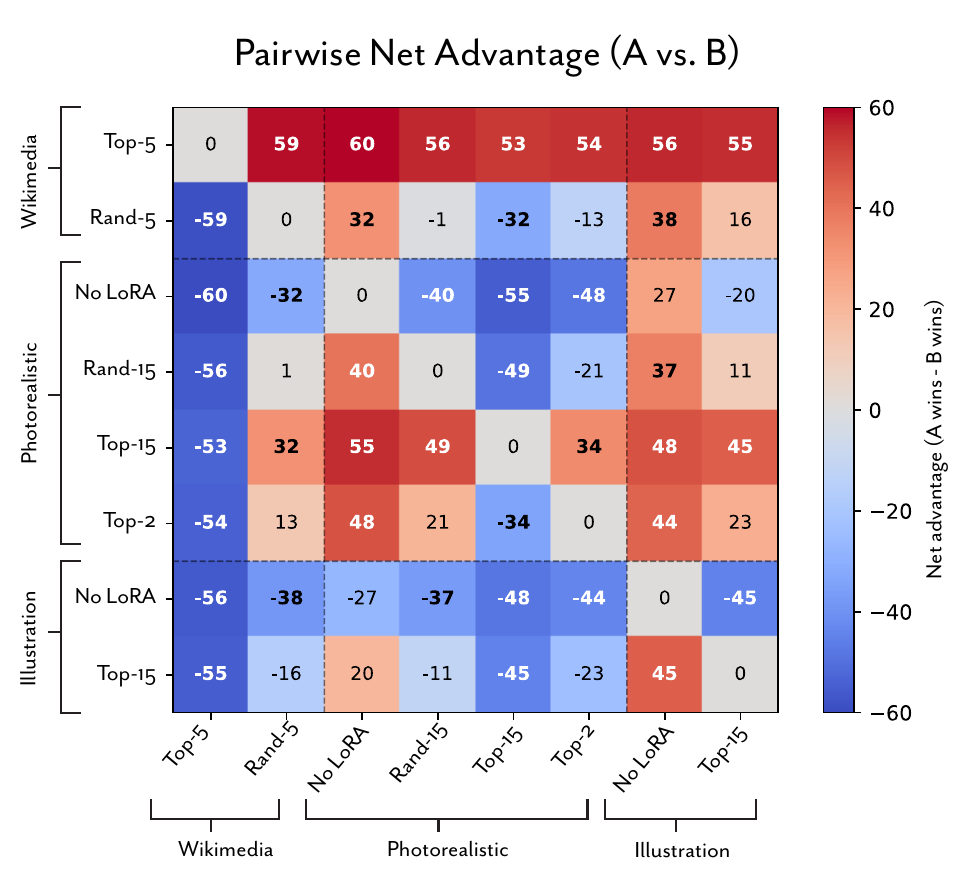}
  \caption{Pairwise net advantage among all conditions.
  Warm (red) colors denote that the row method performs better on more topics.
  LoRA-QuizRank (Top~15) achieves the most consistent advantage. The `Wikipedia' rows/columns reflect the score for real images (non-generated) for the topic in the Wikimedia Commons.}
  \label{fig:net-adv-heatmap}
\end{figure}

Figure~\ref{fig:net-adv-heatmap} presents the net advantage at the topic-level between all conditions.  
Each cell shows how many topics favor the row method over the column method.  
Across 60 topics, \textbf{QZLoRA (Top-15)} exhibits a clear `dominance' (as reflected by the red boxes), 
surpassing almost all other methods on most topics. \textbf{QZLoRA (Top-2)} remains competitive, but is less stable. As expected, \textbf{Top–5 (Real)} serves as a reference upper bound, consistently outperforming all the generated results. Essentially, the best real images of an object will win out over a generated version. However, when looking at a random set of real images (\textbf{Wikipedia Random–5)}, we see a wider variation and which the generative models can beat. Baselines, such as \textbf{No-LoRA}, often score more poorly (as reflected in the blue boxes), indicating limited generalization of the topic and visual accuracy.  Although illustration topics show lower absolute accuracies, \textbf{QZLoRA (Top-15, Illustration)} still outperforms its baseline, demonstrating that QuizRank-guided image selection enhances fine-tuning robustness even in stylistically different domains.

\subsection{Effect of $k$ on Performance}
\begin{figure}[t]
  \centering
  \includegraphics[width=\linewidth]{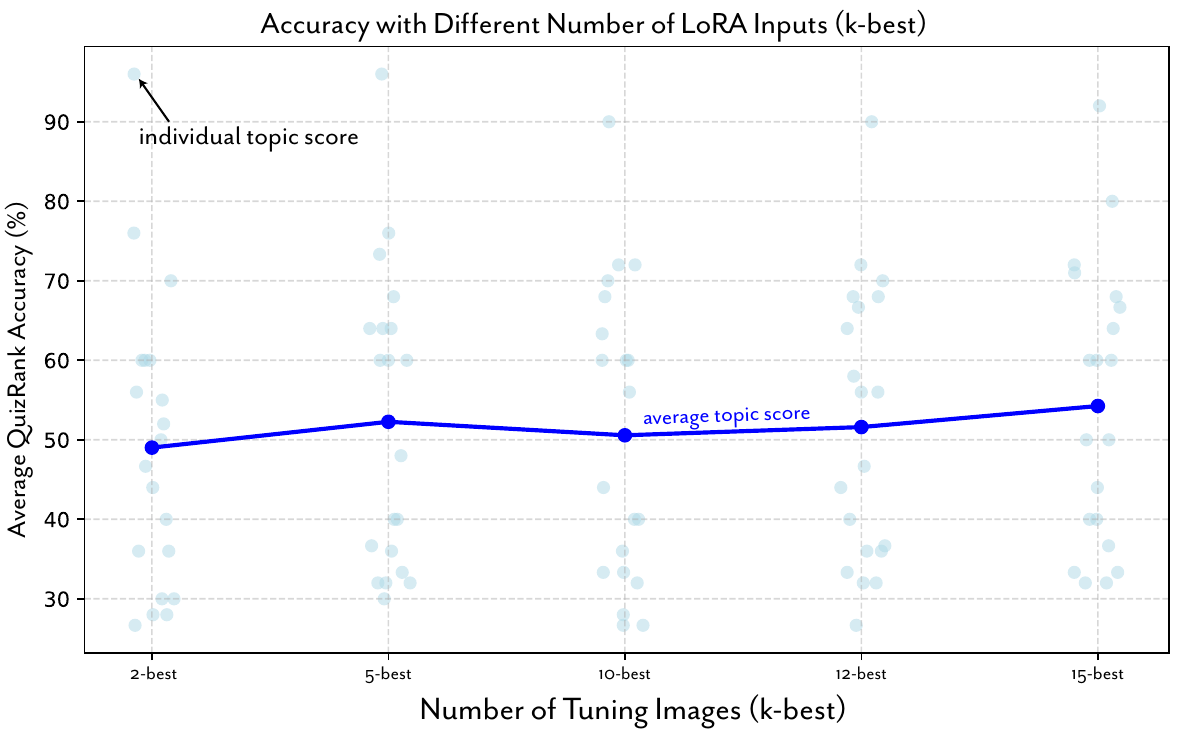}
  \caption{Accuracy change with different numbers of best images selected for LoRA fine-tuning. 
  The results show that using more best images improves the model's accuracy, 
  with the most significant improvement observed when increasing from 5 to 15 images.}
  \label{fig:accuracy_change}
\end{figure}

To investigate the effect of selecting different numbers of top-$k$ images for LoRA fine-tuning, we analyzed the accuracy changes across five conditions: 2, 5, 10, 12, and 15 best images selected for training. Figure~\ref{fig:accuracy_change} shows the accuracy of LoRA fine-tuning using these different levels. Each point represents the average accuracy calculated from 20 randomly topics (of the original 60). The results show that the use of different numbers of images leads to small but consistent improvements in accuracy with increasing $k$. The performance slightly improves from 2 to 5 images, remains relatively stable between 5 and 12, and reaches the highest value when 15 images are used. This pattern suggests that increasing the number of high-quality images generally benefits LoRA fine-tuning, though the overall improvement becomes less pronounced beyond a certain point. 

\subsection{Input-Output Accuracy Correlation}
\begin{figure}[!t]
    \centering
    \includegraphics[width=0.9\linewidth]{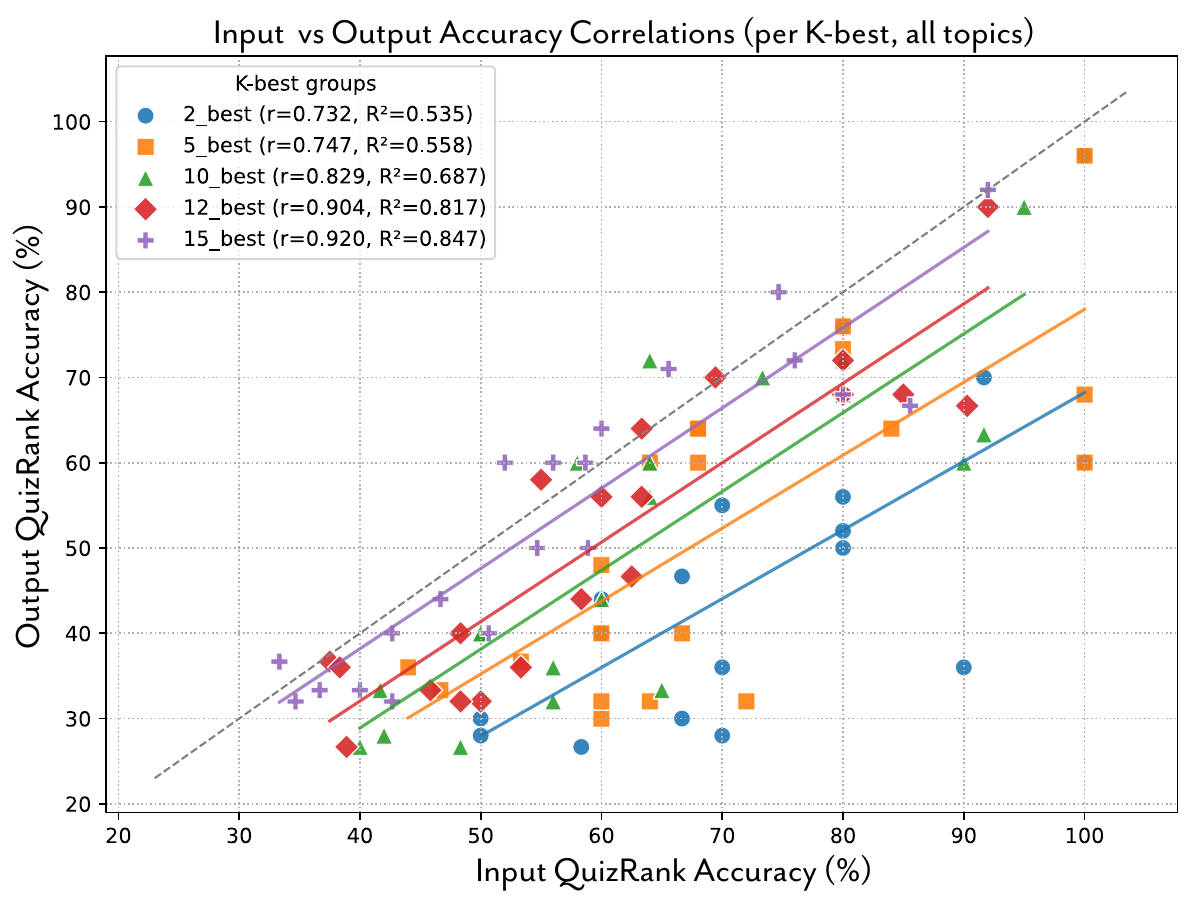}
    \caption{Correlation between input and output image accuracies across different $K$-best groups. Higher correlations at larger $K$ indicate improved stability and transfer quality.}
    \label{fig:correlation}
\end{figure}

To further examine how the quality of the input images influences the generated outputs, we analyzed the correlation between the accuracies of the input and output images in different $k$-best groups (Figure~\ref{fig:correlation}). A clear positive correlation is observed for all $k$ values, showing that higher-quality input images tend to produce higher-quality generated results. 

As $k$ increases, both the correlation coefficient ($r$) and determination coefficient ($R^2$) rise consistently—from $r=0.732$ at $K=2$ to $r=0.920$ at $K=15$. Moreover, the fitted regression lines become steeper and closer to the identity line, suggesting that the fine-tuning process becomes more stable and the output accuracy more directly reflects the input image quality. This indicates that including more high-quality samples not only improves overall accuracy but also improves the reliability of image-to-image knowledge transfer. This result may enable a future optimization where the initial distribution of QuizRank scores for the \textit{real} images may indicate how many are needed for fine-tuning.

\subsection{Effect of Topic Popularity}
We hypothesized that topic popularity--and consequently, more images--may impact generative performance. We would anticipate that the baseline model (without LoRA fine-tuning) would perform better with more images.  As a simple test of this, we analyzed the relationship between  the \textbf{No-LoRA} accuracy and the number of available Wikimedia images per topic. We find that the correlation is quite weak (\textit{Pearson} $r = 0.135$, $p = 0.30$), indicating that topics with more reference images on Wikipedia do not necessarily lead to higher-quality generations.  Thus, more images do not necessarily directly improve the performance of a model. This finding suggests that a model's topic `familiarity' is not directly determined by the volume of public data, but likely by exposure to latent representation during model pretraining. This further indicates that QZLoRA can isolate the best images from potential inputs of various sizes.

\section{Discussion}
The results of our study demonstrate that using QZLoRA can significantly enhance the effectiveness of LoRA fine-tuning. Prioritizing images that are most aligned to key visual properties of objects can improve both the semantic accuracy and visual consistency of the generated images. Compared to random selection, QZLoRA reduces noise and ensures that the model is fine-tuned on the most informative and topic-representative samples. The advantage is particularly noticeable for less common or visually ambiguous topics, where random selection tends to include irrelevant or low-quality images, leading to less coherent results.

The results also highlighted that fine-tuning with a moderate number of top-ranked images yielded the best performance, with diminishing returns observed beyond a certain threshold. It suggests that more images do not necessarily lead to better outcomes, reinforcing the idea that image quality is a more important factor than quantity.

Although effective for a broad set of topics, several limitations should be considered. The effectiveness of QZLoRA still depends on the availability of relevant high-quality images for each topic. For topics with insufficient or low-quality image data, QZLoRA will struggle to select ideal training samples. In such cases, performance may degrade to the level of random image selection. 

Additionally, while we have tried to test across a variety of representative topics, more testing may help clarify QZLoRA's practical limitation. For example, we have observed that QuizRank-based ranking is best on topics that have a `tendency' towards representative visual examples. A topic that is too broad--for example, the thrush bird family, which has 193 species--may not have representative images. An extension of QuizRank and QZLoRA may be to identify clusters of images based on \textit{which} questions an image helps answer correctly and training different models for each cluster. Second, for topics that are an instance (e.g., Donatello's David or the Newgrange monument) rather than class (e.g., a Mountain Bluebird) may be more challenging for generative models (see Figure~\ref{fig:teaser_realistic}). Although QZLoRA outperforms other approaches, it can still hallucinate aspects of the object.

Additionally, our experiments focus on the use of QZLoRA for a particular diffusion model (1.5). While we expect benefits to transfer to other models, additional testing would be necessary. However, the advantage of using QuizRank to evaluate generative outputs is that it allow for more direct generative comparison. 

Practically, the use of QZLoRA incurs some computational cost (i.e., generating the questions and VLM judgement, see~\cite{ji2025quizrankpickingimagesquizzing}). However, our experience has been that this cost is relatively low compared to the actual LoRA procedure. Finally, while QZLoRA is promising for generating non-photorealistic images, additional evaluation is needed for different illustration types. A limitation of most images in our testing is that the inputs are photorealistic. A possible future approach is to apply a style transfer algorithm or filter to the input images before fine-tuning. 

\section{Conclusion}
In the work, we demonstrate the value of the QZLoRA approach in optimizing fine-tuning by selecting high-quality images that visually represent a target topic. Through our approach, we showed that focusing on the relevance of training data, rather than sheer quantity, leads to more reliable and coherent image generations. By incorporating the idea of `image informativeness', we ensured that only the most useful and visually aligned images were used for fine-tuning, improving both accuracy and stability across diverse topics. We also demonstrate how QuizRank can be used to objectively contrast the output of different image generation approaches. This approach may help improve the evaluation of new generative models and approaches.

{
    \small
    \bibliographystyle{ieeenat_fullname}
    \bibliography{custom}
}

\appendix

\section{Appendix}

\subsection{Generative Prompts}
Generative prompts used the description (top sentence) from Wikipedia for each topic. All prompts followed this pattern. For example, for generating photorealistic images of the Gujia, the prompt was:

\begin{lstlisting}[basicstyle=\footnotesize]
Positive Prompt:
Generate the image of Gujhia (also known as gujiya, gujia, gughara, pedakiya, purukiya, karanji, kajjikayalu, somas, or karjikayi), a sweet deep-fried pastry popular in the Indian subcontinent, realistic food photography, high resolution, detailed textures, natural lighting, shallow depth of field, several gujhias arranged neatly on a red plate, stainless steel or ceramic red plate.

Negative Prompt:
illustration, drawing, painting, vector art, cartoon, flat colors, low quality, blurry, CGI, 3D render, fake texture, plastic look, overexposed, underexposed, watermark, text
\end{lstlisting}

For illustration style images we used:

\begin{lstlisting}[basicstyle=\footnotesize]
Positive Prompt:
Generate the image of Gujhia (also known as gujiya, gujia, gughara, pedakiya, purukiya, karanji, kajjikayalu, somas, or karjikayi), a sweet deep-fried pastry popular in the Indian subcontinent, vector illustration, flat colors, bold clean lines, simplified texture, smooth shading, 2D drawing, food illustration style, minimalistic background.

Negative Prompt:
realistic, photorealistic, photo, natural lighting, shadows, depth of field, glossy surface, crisp texture, CGI, 3D render, high contrast, overexposed, underexposed, watermark, text
\end{lstlisting}

\end{document}